\documentclass[runningheads]{llncs}
\usepackage{graphicx}
\usepackage{algorithm2e}
\usepackage{amsmath}
\usepackage{bm}
\usepackage{bbm}
\usepackage{amsfonts}
\usepackage{caption}
\usepackage{subcaption}
\usepackage{hyperref}
\usepackage{adjustbox}
\begin{document}
\title{Encoding Categorical Variables with Conjugate Bayesian Models for WeWork Lead Scoring Engine\thanks{Supported by WeWork Companies, Inc.}}
\titlerunning{CBM Encoding}
%
%
\author{Austin Slakey\inst{1} \and
Daniel Salas\inst{1} \and
Yoni Schamroth\inst{1}}
%
%
\institute{WeWork Companies, Inc., New York, NY 10011, USA\\
\email{\{austin.slakey,daniel.salas,yoni.schamroth\}@wework.com}\\
\url{http://www.wework.com}}
\maketitle              
\begin{abstract}
Applied Data Scientists throughout various industries are commonly faced with the challenging task of encoding high-cardinality categorical features into digestible inputs for machine learning algorithms. This paper describes a Bayesian encoding technique developed for WeWork's lead scoring engine which outputs the probability of a person touring one of our office spaces based on interaction, enrichment, and geospatial data. We present a paradigm for ensemble modeling which mitigates the need to build complicated preprocessing and encoding schemes for categorical variables. In particular, domain-specific conjugate Bayesian models are employed as base learners for features in a stacked ensemble model. For each column of a categorical feature matrix we fit a problem-specific prior distribution, for example, the Beta distribution for a binary classification problem. In order to analytically derive the moments of the posterior distribution, we update the prior with the conjugate likelihood of the corresponding target variable for each unique value of the given categorical feature. This function of column and value encodes the categorical feature matrix so that the final learner in the ensemble model ingests low-dimensional numerical input.  Experimental results on both curated and real world datasets demonstrate impressive accuracy and computational efficiency on a variety of problem archetypes. Particularly, for the lead scoring engine at WeWork -- where some categorical features have as many as 300,000 levels -- we have seen an AUC improvement from 0.87 to 0.97 through implementing conjugate Bayesian model encoding.
\keywords{Categorical variables  \and Conjugate priors \and Ensemble models.}
\end{abstract}
\section{Introduction}
Lead scoring at WeWork involves calculating the probability of a potential new member to book a tour at one of our locations. There is significant business impact when sales associates use these scores to decide who and when to contact. We prioritize high probability leads with product-specific sales specialists while routing low probability leads to less costly, automated sales approaches. The classification algorithms that we deploy for this task must ingest numerical data. Thus, one of the key challenges we face in model development is encoding non-continuous, categorical features which have no intrinsic order and many unique values into meaningful numerical input. Lead scoring and similar projects demand a solution to handle categorical data that balances the complexity trade-off of encoding techniques while enabling a model to train quickly, score in real time, and perform well on rare and unseen categories. The resulting model should be able to generalize, for example, to a lead associated with a rarely seen company and interested in brand new locations.

In real world databases, the cardinality, or number of distinct values, of a column is routinely high for a variety of reasons. Geographical locations, retail products, or surgical procedures, for example, will vary due to the sheer vastness of possibilities.  Form input, on the other hand, may vary due to user error (typographical, encoding, or special character errors) and diverse morphological representations of the same entity (synonyms, aliases, or abbreviations). At WeWork, we encounter many of these common issues. For instance, at the top of our sales funnel, potential customers enter into our system as ``leads'' through diverse sources (website pages, social media, search engines, broker referrals, employee referrals, etc.) from any location around the globe.  We then match the lead to their respective companies in order to join in details such as industry, headquarter location, and company size.  Further still, sales associates and the leads themselves can manually add details such as interested locations and number of desks via form input. The resulting lead object contains many unique values. Below is the cardinality for several commonly used features at WeWork:

\begin{table}
\centering
\caption{Cardinality of categorical features in Lead Scoring dataset.}\label{tab1}
\begin{tabular}{|p{6cm}|p{4cm}|}
\hline
Column &  Cardinality\\
\hline
{\bfseries Source Detail} &  18,162\\
{\bfseries Locations Interested} &  29,641\\
{\bfseries Industry} &  36,085\\
{\bfseries Email Domain (proxy for company)} &  346,727\\
\hline
\end{tabular}
\end{table}

These high-cardinality categorical columns are often critical features in machine learning tasks such as classification and regression.  However, there exists no straightforward engineering or statistical methods to handle them.  The most common method of encoding categorical variables for machine learning algorithms is to one-hot encode them by creating a binary column for each unique value of the categorical column. At WeWork that equates to 36,085 dimensions to encode industry alone. Furthermore, many of these levels are rarely seen yielding a sparse matrix -- one with a high proportion of zeros. This naive approach requires intense computational capacity, neglects previously unseen values, and worse, encounters every downfall associated with the curse of dimensionality including combinatorial explosion, counterintuitive geometric phenomena, sampling from high dimensional space, and issues with convergence of optimization methods \cite{ref_1,ref_2}. 

Preprocessing and standardizing the column by clustering or combining like values is one solution, but valuable information can be lost when the merging methodology is not exposed to statistical methods \cite{ref_3}. More sophisticated encoding techniques such as autoencoders and, more recently, word2vec introduced by Mikolov et al. perform extremely well and are the focus of much current research \cite{ref_4,ref_5}. Although quite promising for machine learning tasks, presently, these techniques require significant training time and careful implementation which can be prohibitive for use in rapid model development.

To encode high-cardinality categorical variables, we introduce a technique based on traditional Bayesian statistics. This technique is a paradigm for ensemble modeling, specifically stacking, where the base learner consists of a problem-specific conjugate Bayesian model (CBM). A conjugate prior model follows from Bayes' Theorem whereby the probability of a certain event occurring is related to prior knowledge. For data $y$ and parameter $\theta$, Bayes' Theorem states that the posterior distribution is the normalized product of the likelihood and the prior distribution:

\begin{equation}
p(\theta|y) = \frac{p(y|\theta)p(\theta)}{p(y)}
\end{equation}

In this equation $p(\theta)$ is the prior for $\theta$, $p(y|\theta)$ is the likelihood of data $y$ given $\theta$, and $p(\theta|y)$ is the posterior distribution of parameter $\theta$ according to data $y$.  A conjugate Bayesian model, in particular, is one in which the prior and posterior are of the same distribution family \cite{ref_6}. These conjugate Bayesian models are analytically derived, efficiently computed, and commonly used as standalone, theoretically sound statistical models for prediction tasks. In this paper's proposed technique, we encode categorical features with moments of the posterior distributions fit on each unique value through Bayes' theorem. The final learning algorithm in the ensemble model can improve upon the simple conjugate Bayesian models by accounting for variance of posterior distributions and interactions among other features.

For the lead scoring engine at WeWork, which tackles a binary classification problem, we implement CBM encoding with a Beta-Binomial model. The encoder is initialized with a prior Beta distribution, and we update with the conjugate Binomial likelihood for each unique value of a given categorical variable. The categorical variable is then encoded via a function of column and value that yields the first two moments -- mean and variance -- of the posterior Beta distribution. Thus, each element in a categorical feature matrix is represented in 2 dimensions. After implementing this at WeWork for our lead scoring engine, we experienced impressive improvements in accuracy over our previous production model: Area Under the Receiver Operating Characteristic curve (AUC) improved from 0.87 to 0.97.  
 
We present experimental results for this technique on a variety of datasets.  On six publicly available, benchmarked datasets, results indicate that CBM Encoding performs as well or better than other common encoding techniques in regards to accuracy and outperforms most in computation time. We also present results on two real world datasets for which some common encoding techniques such as one-hot encoding are no longer feasible due to extreme high-cardinality. In these practical settings, the great reduction in training time by using CBM Encoding allows for increased resources on experimentation with model prototyping, feature engineering, and tuning of the final learning algorithm. The promising results suggest further research into optimizing this technique such as choosing an optimal prior or taking into account similarity across levels of a categorical feature.

\section{Related Works}
Extensive work and research has been conducted in incorporating high-cardinality categorical features into machine learning applications. Multiple solutions have been proposed each dealing with different aspects of the challenge of dealing with this class variables.

Traditionally, handling a categorical variable with $n$ levels meant introducing $n-1$ binary dummy variables into the feature set, a method known as dummy-encoding. The popular One-Hot Encoding is an extension of this method to $n$ dichotomous variables where each level is represented by a vector of zeros with an entry of $1$ each time that particular level appears in the data \cite{ref_12}. As mentioned previously, applying these methods on variables of high cardinality suffers from a number of shortcomings: (1) The curse of dimensionality - The need to introduce an additional field for each distinct level results in a massive increase in the dimension of the training data being used. (2) Consequently, the computational power needed to deal with such large volumes of data become prohibitive. (3) Such variables generally also have very skewed distributions with most of the levels hardly appearing in the data. Since these methods essentially treat each individual level the same, they fall short in handling sparsely represented entries. (4) Finally these methods do not fare well in the presence of new and unseen levels. There is likewise no inherent support for missing values.

Weinberger, K. et al.  propose a method for dimensionality reduction which they call the  hashing trick \cite{ref_10}. In this approach hashing techniques are used to map the feature space to a reduced vector space. This has the desirable effect of not only reducing dimensionality and therefore computation, but it also provides an elegant way of dealing with new or missing entries. One shortcoming is the presence of collision where different entries are assigned the same hash. However it has been shown the the gain in accuracy and performance outweigh the effects of this phenomenon. 

Another approach used to reduce the dimensionality of the problem is to cluster similar levels together. This not only reduces the cardinality of the category in question, but as a consequence also increases the frequency of each level reducing sparsity. Various measures of similarity can be used to achieve this clustering. A basic technique would be to group all low frequency entries into one bucket. Though simple to implement, this approach has many shortcomings and generally results in major information loss and reduction of predictive power.

Similarity might also be quantified by measuring the morphological resemblance of the actual text describing the level itself. Patricio Cerda et al. propose such a similarity encoding technique to encode ‘dirty’, non-curated categorical data \cite{ref_3}. This method is shown to outperform classic encoding techniques in a setting of high redundancy and very high cardinality.

Finally similarity might be measured in relation to the actual target response variable itself. Contrast encoding, for instance, builds upon classic statistical contrast techniques often used in analysis of variance or regression to encode each level of the variable.  In this fashion nominal entries of the categorical variable are replaced with continuous values which measure its relation to some previously calculated statistic. For example the helmert contrast compares the mean of the target variable for a particular level with the mean of the remaining levels. Contrast sum compares the mean of each level to the overall mean. Other contrasts include backward / forward difference, polynomial and leave-one-out \cite{ref_13,ref_14}.

These methods of continuousification of a nominal variable can be extended to more advanced preprocessing schemes which take full advantage of the target variable at hand. Micci-Barreca apply an Empirical Bayes approach to estimate the mean of the target variable, $Y$, conditional on each level of the categorical, $X$, at hand, e.g. $E[Y=y_{i}|X=x_{i}]$ \cite{ref_7}. Conversely the MDV approach encodes each level by measuring the mean of the respective level conditional on the target variable $E[X=x_{i}|Y=y_{i}]$ \cite{ref_15}. Both these methods are variations of the more general Value Difference Metric (VDM) continuousification scheme \cite{ref_16,ref_17}. Once the entries have been transformed on a continuous scale further clustering can be performed grouping similar levels together \cite{ref_7}.

In our proposed method, we build on the approaches outlined above while drawing further inspiration from the works of Vilnis and McCallum where they introduce the novel idea of using gaussian embeddings to learn word representations \cite{ref_18}. Word embedding is a method of mapping a particular word into a N-dimensional vector of real numbers. Embeddings have the ability to translate large sparse vectors into a lower-dimensional space whilst preserving semantic relationships and was popularized by Mikolov, Tomas, et al. in their paper on `word2vec' \cite{ref_5}. Gaussian embedding moves beyond the point vector representation, embedding words directly as Gaussian distributional potential functions and distances between these distributions are calculated using KL - divergence. In this `word2gauss' approach, a preprocessing step is similarly required to learn the mean and variance of each distribution, thus the uncertainty surrounding the word is fully captured and utilized.

\section{Methodology}
To address the issue of encoding categorical variables in settings with high cardinality, we formally introduce the CBM encoding model. Through standard conjugate Bayesian models we are able to represent categorical features accurately and in low dimensions.

\subsubsection{Notation}
 In the following sections we write matrices with a bold letter $\mathbf{X}$, elements of a matrix in the $n$-th row and $m$-th column as $\mathbf{X}_{nm}$, and set notation with braces $\{\mathbf{X}_{nm}\}_n$ denoting the unique values over all rows $n$ for column $m$.
 
 \subsection{CBM Encoding} To define a machine learning problem in the context of categorical data, let $\mathbf{X}$ be an $N \times M$ matrix with row vectors $X_n$ and column vectors $X^T_m$. Let $y$ be an $N$-dimensional vector such that $y_n$ is the observed value corresponding to $X_n$. Then $D = (\mathbf{X}, y)$ is a dataset where $D_n \equiv (X_n, y_n)$ is the $n$-th observation. 
 
 In context of the categorical data problem, column $X^T_m$ with cardinality $K_m$ has a domain $V(m)=\{\mathbf{X}_{nm}\}_n$ containing unique nominal values $v \in 1,...,K_m$. Through CBM Encoding, we attempt to learn probabilistic models for $\mathbf{X}$ then represent the categorical features from $\mathbf{X}$ via $Q$ moments from the corresponding learned probabilistic models. The resulting $N \times QM$ matrix is $\mathbf{Z} \in \mathbb{R}$. Thus, we can instead model $\hat{D} = (\mathbf{Z}, y)$ which is a simpler problem in general.
 
 The model presented in this paper consists of two layers. First, the Local Layer builds a na\"ive model for each feature based on the likelihood of observing the target variable for each value of the feature. Then, the Encoding Layer generates a distributed representation of the categorical features from the corresponding probabilistic models created in the Local Layer.


\subsubsection{The Local Layer.}

For each categorical column $X^T_m \in 1,...,M$, the Local Layer builds probabilistic model $\mathcal{L}(\theta_{mv}|y)$ of the likelihood of the target variable $y$ for each nominal value $v \in V(m)$ with a distribution parameter vector $\theta_{mv}$. The Local Layer first defines a prior distribution $p(\theta_{mv})$ for each model, and then computes the posterior distribution $p(\theta_{mv}|y) \propto \mathcal{L}(\theta_{mv}|y)p(\theta_{mv}).$

In practice, we can set $p(\theta_{mv})$ to be the same distribution for all values of $m$ and $v \in V(m)$ if we have no prior knowledge of the parameter. Alternatively, we could use expert knowledge if, for example, one expects differing behavior for rare categories or similarity between certain nominal values. Second, we note that each parameter $\theta_{mv}$ is defined separately for each value of $m$ and $v \in V(m)$. As a result, we have that $\mathcal{L}(\theta_{mv}|y) = \mathcal{L}(\theta_{mv}|\hat{y}_{mv})$ where $\hat{y}_{mv} \equiv \{y_n\ : \mathbf{X}_{nm} = v\}_n$.

\subsubsection{The Encoding Layer.}
Building from the na\"ive probabilistic models created in the Local Layer, the Encoding Layer saves the computation of the posterior's first $Q$ moments as a function of column and value, $f(m,v)$. This allows us to efficiently represent any element $X_{mn}$ in $Q$ dimensions. Therefore, we can build matrix $\mathbf{Z}$ in a straightforward way by computing $f(\mathbf{X}_{nm},m) \forall n,m$. Specifically, for $q$ in $1,...,Q$ we assign the $q$-th element of $f(m,v)$ to $\mathbf{Z}$ in the following manner:
\begin{equation}
\mathbf{Z}_{n,(m-1)Q+q}\leftarrow{f(\mathbf{X}_{nm},m)_q}
\end{equation}

In our implementation, we take advantage of conjugate Bayesian models in order to analytically derive the posterior distributions of our parameters exactly. However, we only rely on the moments of the posterior distributions, so it would be possible to extend this implementation to non-conjugate models by obtaining estimates of the moments using approximate Bayesian inference methods like MCMC or variational methods.

The Local Layer and Encoding Layer together comprise CBM Encoding. The full algorithm is outlined in the Appendix in Section 6.1. Specific implementations for binary classification, multiclass classification and regression are explicitly defined in Table \ref{tab:cbm}.










\section{Experiments}
In order to empirically evaluate the performance of encoding categorical features with conjugate Bayesian models, we set up two experiments.  The first experiment compares performance against four common encoders in scenarios with various learning algorithms.  The datasets for this experiment are readily available and have established benchmarks for performance.  The second experiment focuses explicitly on high cardinality categorical data and the CBM Encoder's performance against other commonly evoked techniques as dimensionality and sparsity increase.  Descriptions and links to all datasets are provided in the Appendix.

To preserve a reproducible environment, all experiments utilize the standard encoders and packages available in the Python module scikit-learn and the scikit-learn contributed Category Encoders package.  Default parameters for all encoders and algorithms are used except for increasing the number of estimators for Random Forest to 100.  Note that non-categorical features are normalized using scikit-learn's StandardScaler class. Lastly, we utilize Docker to maintain consistent requirements and to host experiments on a server.  All code and data along with the Docker File to run are available on GitHub \footnote{GitHub: \url{https://github.com/aslakey/CBM_Encoding}.}.

\subsection{Conjugate Bayesian Models and Prior Initialization}
For the following experiments we use Binomial, Multinomial, and Gaussian likelihoods for binary classification, multiclass classification, and regression, respectively. The formulations for these conjugate Bayesian models are listed in Table \ref{tab:cbm}. Notation in the table follows standard conventions. For all conjugate Bayesian models, $n$ is the number of instances in the dataset. For the Beta-Binomial in particular, target variable $y$ can take on values \{0,1\}. For the Dirichlet-Multinomial, target variable $y$ can take on multiple values, specifically $y_i\in{(1,...,K)}$. Therefore, parameter $\bm{\alpha}$ is an array of length $K$. For the Normal-Inverse Gamma-Gaussian model, $\bar{x}$ is the sample mean and $\hat{\sigma}$ is the sample variance.

\begin{table}
\caption{Conjugate Bayesian Models used in Experiments 1 and 2}
\label{tab:cbm}
\begin{adjustbox}{width=1\textwidth}
\small
\begin{tabular}{|l|c|c|p{4cm}|}
\hline
Prior & Prior Parameters & Likelihood & Posterior Parameters\\
\hline
{\bfseries Beta} &  $\alpha, \beta$  & {\bfseries Binomial} & $\alpha+\sum_{i=1}^{N}{{y_i}}, \beta +n-\sum_{i=1}^{n}{{y_i}}$\\
{\bfseries Dirichlet} &  $\bm{\alpha}$ & {\bfseries Multinomial} & $\alpha_k+\sum_{i=1}^{n}{\mathbbm{1}_{k}{(y_i)}}$\\
{\bfseries Normal-Inverse Gamma} &  $\mu,\nu,\alpha,\beta$ & {\bfseries Gaussian} & $\frac{\nu\mu+n\bar{x}}{\nu+n},\nu+n,\alpha+n/2,\beta+\frac{1}{2}n\hat{\sigma}^{2}+(\frac{n\nu}{\nu+n})(\frac{(\bar{x}-\mu)^{2}}{2})$\\
\hline
\end{tabular}
\end{adjustbox}
\end{table}

In the following experiments, priors for all Beta-Binomial models are initialized using $y$ from the training set so that $\alpha = \frac{\sum_{i=1}^{n}y_i}{n}$ and $\beta=1-\alpha$. Similarly, for the Dirichlet-Multinomial model, we let $\alpha_k=\sum_{i=1}^{n}\mathbbm{1}_{k}{(y_i)}$ then normalize: $\alpha_k \gets \frac{\alpha_k}{\sum_{k=1}^{K}{\alpha_k}}$. Finally, for the Normal-Inverse Gamma-Gaussian model, we initialize $\mu$ with the sample mean, $\bar{y}$, from the training data and $\beta$ with the sample variance, $\hat{\sigma}^{2}$.

\subsection{Comparing Encoders}
\subsubsection{Overview.} The first experiment is designed to compare CBM Encoding against common encoders on benchmark datasets.  We selected three problem types and demonstrated results on two datasets for each problem. The comparison encoders are the following for all problem types and datasets: one-hot, binary, ordinal, and target. The learning algorithms employed include Gradient Boosting Trees, Logistic Regression, Multi-Layer Perceptron, Random Forest, and Ridge Regression.  Reported results are accuracy for binary and multiclass Classification and $r^2$ for regression from 10-fold cross validation with a corresponding error bar of +/- one standard deviation. In the following figures, a subscript of $\it{(m)}$ indicates that the CBM Encoder's function of column and value yields the first moment, or mean, of the the posterior distribution, and $\it{(mv)}$ indicates both mean and variance.
\subsubsection{Results.} 
\begin{figure}[t]
  \begin{subfigure}{0.49\textwidth}
    \includegraphics[width=\textwidth]{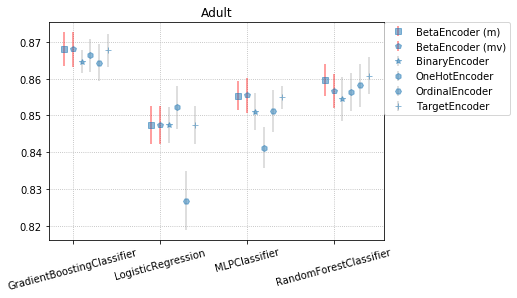}
    \caption{Adult}
  \end{subfigure}
  \begin{subfigure}{0.49\textwidth}
    \includegraphics[width=\textwidth]{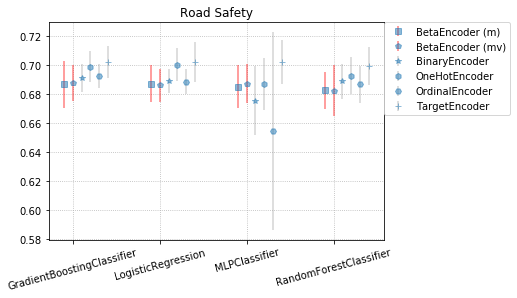}
    \caption{Road Safety}
  \end{subfigure}
  \caption{Comparing Beta Encoder against common encoders for tasks in binary classification}
  \label{fig:binary}
\end{figure}

The first problem type is binary Classification which is the task of assigning an element to one of two classes. The CBM Encoder for this problem type uses the Beta-Binomial model, resulting in dimensionality of $1*Q$ for $Q$ chosen moments. Experimental results, displayed in Figure~\ref{fig:binary}, indicate remarkable accuracy for the Beta Encoder compared with other encoders for the Adult dataset. It is worth noting the similar performance to Target Encoding as the mathematical formulations are quite similar to CBM Encoding when using a Beta-Binomial model.

In the Road Safety dataset, however, the Beta encoder lags slightly in performance across most of the learning models. This particular dataset was invoked in the 2018 paper on Similarity Encoding by Cerda et al. because of its unique characteristics \cite{ref_3}. The task is to predict gender from make and model of a car; however, most make and models are rarely seen across both training and testing datasets. Hence, Similarity encoding performs quite well by grouping semantically similar makes and models together. CBM and Target encoding, however, tend to overfit the training set in such scenarios. For example, with a simple Logistic Regression meta learning algorithm, the average accuracy on training data with CBM Encoding significantly outpaced the out of fold accuracy at 0.854 compared to 0.690.  The common workaround invoked for Target encoding is application of Gaussian noise (default in the Category Encoders package). For CBM Encoding, the same workaround can be utilized along with initializing a stronger prior or optimizing sampling methods of training data.

\begin{figure}[t]
  \begin{subfigure}{0.49\textwidth}
    \includegraphics[width=\textwidth]{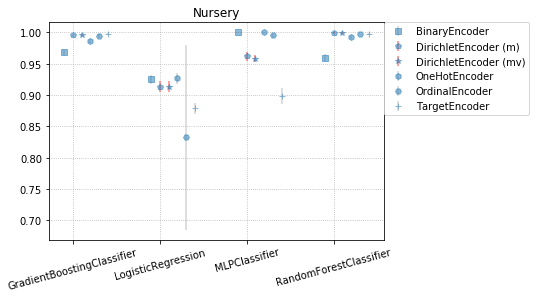}
    \subcaption{Nursery}
  \end{subfigure}
  \begin{subfigure}{0.49\textwidth}
    \includegraphics[width=\textwidth]{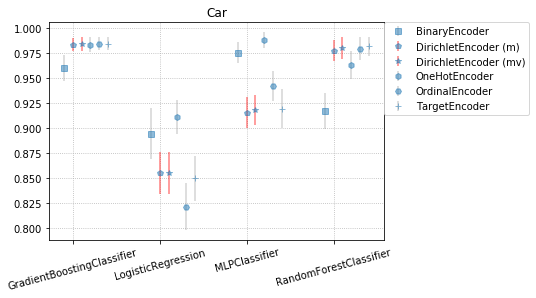}
    \subcaption{Car Evaluation}
  \end{subfigure}
  \caption{Comparing Dirichlet Encoder against common encoders for tasks in Multiclass Classification}
  \label{fig:multiclass}
\end{figure}

For multiclass classification problems, the conjugate Bayesian model employed is Dirichlet-Multinomial, a multivariate extension of the Beta-Binomial. For a problem with $K$ possible classes, the encoder results in $K*Q$ dimensions for $Q$ chosen moments. In this case, CBM Encoding notably differs from Target encoding which has no intuitive probabilistic interpretation in scenarios where target classes have no intrinsic order. Results from multiclass classification problems (Figure~\ref{fig:multiclass}) show particularly good performance for CBM Encoding with tree based algorithms such as Gradient Boosting and Random Forest. The results are intuitive because the meta learning algorithm can improve upon the base predictions of the Dirichlet-Multinomial model by modeling feature interactions.

\begin{figure}[t]
  \begin{subfigure}[b]{0.49\textwidth}
    \includegraphics[width=\textwidth]{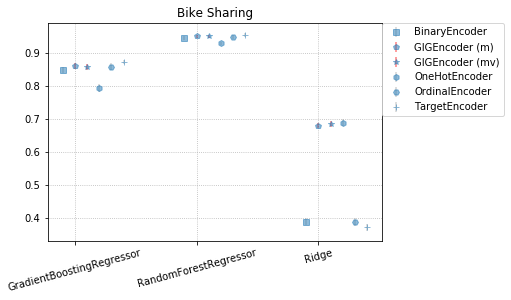}
    \subcaption{Bike Sharing}
  \end{subfigure}
  \begin{subfigure}[b]{0.49\textwidth}
    \includegraphics[width=\textwidth]{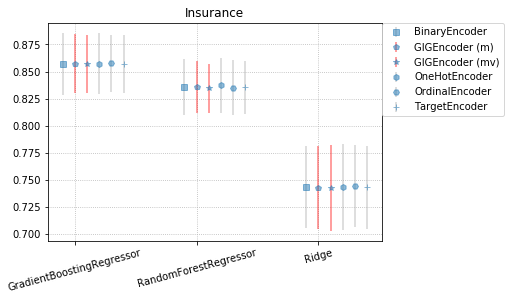}
    \subcaption{Insurance}
  \end{subfigure}
  \caption{Comparing GIG Encoder against other common encoders for tasks in Regression}
  \label{fig:regression}
\end{figure}

In regression problems, we utilize a Normal-Inverse Gamma prior and Gaussian likelihood for CBM Encoding. Because both the mean and variance are modeled, the encoder results in $2*Q$ dimensions for $Q$ chosen moments. Nicknamed the GIG (Gaussian Inverse Gamma) encoder, this technique performs particularly well on regression problems especially on real world datasets with high cardinality and larger sample sizes such as Bike Sharing and PetFinder in Figure~\ref{fig:regression} and Table~\ref{tab:petfinder}, respectively.

\subsection{CBM Encoding in High Cardinality}

The purpose of the second experiment is to test CBM Encoding in situations with high cardinality when standard one-hot encoding is no longer a viable option. We present results on two datasets, Lead Scoring and PetFinder, with comparisons against hashing and one-hot with a $\it{truncator}$ -- grouping categories with a count below some defined threshold. To save on computation time, results are reported on a randomly selected 30\% hold out set. The learning algorithm for all experiments is eXtreme Gradient Boosting (XGB) using default parameters and accessed through the scikit-learn API. For both datasets, we report accuracy and training time which includes fitting of the encoder and fitting of the XGB model. In both experiments, CBM Encoding performs remarkably well from computation and accuracy perspectives, proving that this technique is an attractive option for real world datasets with high cardinality.
\subsubsection{PetFinder.} 
The first results are reported on PetFinder, a dataset from a Kaggle\footnote{Kaggle: \url{https://www.kaggle.com/}.} Data Science competition with the task of predicting how quickly an animal will be adopted. Speed of adoption is defined in discrete classes, so the competition -- and the reported results -- measures accuracy by a weighted Cohen's quadratic weighted kappa score\cite{ref_19}. As comparison encoders, we use hashing with 1,000 dimensions and a one-hot with truncation threshold at 25, i.e. any categories with less than 25 samples in the training set are grouped. For CBM Encoding we tested the Dirichlet-Multinomial model, and because the target classes are ordinal, the GIG encoder.

\subsubsection{Lead Scoring.} The second results are reported on a sub-sample of rows and features from WeWork's own Lead Scoring dataset. Specifically, we selected a random sample of length 256,000 and used five of the most informative features -- three categorical and two numerical. As comparison encoders, we use hashing with 1,000 dimensions and one-hot with truncation threshold at 150.

\subsubsection{Results.} Results in Tables \ref{tab:petfinder} and \ref{tab:leadscoring}, show that CBM Encoding outperforms in accuracy and significantly outperforms in computation time -- 100 times faster than the scikit-learn Hashing Encoder implementation for Lead Scoring. Memory and computation time were significant issues with larger sample sizes for all scikit-learn implemented encoders. Contrarily, CBM Encoding shows only marginal gains in computation time as sample size grows due to the much lower, constant dimensionality. 

CBM encoding with a Beta prior is significantly more accurate for Lead Scoring. On the PetFinder dataset, accuracy measures are fairly high for both conjugate Bayesian models, but the Normal-Inverse Gamma prior and Gaussian likelihood models the Speed to Adoption variable more reliably.
\begin{table}[t]
\caption{Experiment 2. PetFinder}
\label{tab:petfinder}
\centering
\begin{tabular}{|p{3.5cm}|p{2cm}|p{2cm}|p{3cm}|}
\hline
Encoder &  Dimensions & Accuracy & Training Time\\
\hline
{\bfseries GIG Encoder} &  32 & 0.3999 & 9.195\\
{\bfseries Hashing Encoder} &  1019 & 0.3777 & 178.0\\
{\bfseries Dirichlet Encoder} &  68 & 0.3771 & 18.26\\
{\bfseries One-Hot Encoder} &  355 & 0.3485 & 59.93\\
\hline
\end{tabular}
\end{table}

\begin{table}[t]
\caption{Experiment 2. Lead Scoring}
\label{tab:leadscoring}
\centering
\begin{tabular}{|p{3.5cm}|p{2cm}|p{2cm}|p{3cm}|}
\hline
Encoder &  Dimensions & Accuracy & Training Time\\
\hline
{\bfseries Beta Encoder} &  5 & 0.9253 & 7.660\\
{\bfseries Hashing Encoder} &  1002 & 0.9152 & 719.9\\
{\bfseries One-Hot Encoder} &  557 & 0.9162 & 302.6\\
\hline
\end{tabular}
\end{table}

\begin{figure}[t]
  \begin{subfigure}[b]{0.49\textwidth}
    \includegraphics[width=\textwidth]{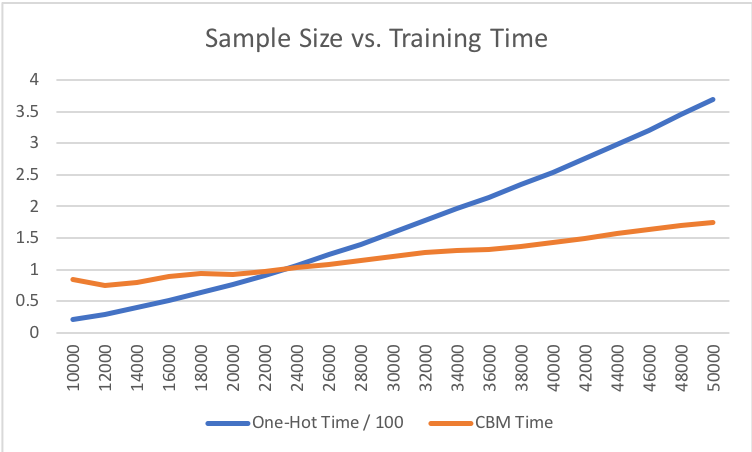}
    \subcaption{Training Time}
  \end{subfigure}
  \begin{subfigure}[b]{0.49\textwidth}
    \includegraphics[width=\textwidth]{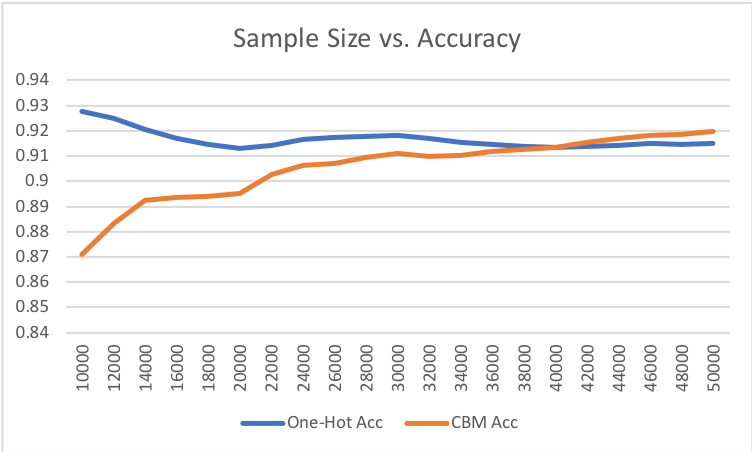}
    \subcaption{Accuracy}
  \end{subfigure}
  \caption{Comparing One-Hot encoding and CBM Encoding as the sample size scales in a real world data set. Measures (time and accuracy) are reported via moving averages over the last five sample sizes to smooth results of XGB's stochastic training. Note that training time for the One-Hot Encoder is scaled by a factor of $1/100$.}
  \label{fig:lead_scoring}
\end{figure}

We specifically tested accuracy and training time at increasing sample sizes in increments of 2,000 for the Lead Scoring dataset with both CBM and One-Hot Encoding. The results in Figure \ref{fig:lead_scoring} are remarkable, yet intuitive. Training with both encodings scale linearly with the sample size, but at much different rates. Because the dimensionality remains constant for CBM encoding, the larger sample sizes result in minimal increase computation time. Training time using CBM encoding increased just over one second from 2,000 to 50,000 samples. However, we experienced significant increases in accuracy as the simple conjugate Bayesian models became more certain and the meta learner could generalize to more nuanced cases. In contrast, training time with One-Hot Encoding increased from 10 seconds to over 400 seconds. The computation time for One-Hot is impacted by the correlation between cardinality and sample size -- more unique values are seen. Furthermore, the One-Hot model became less accurate with increasing sample size which follows from the curse of dimensionality's negative side effects.

\section{Conclusions}
CBM Encoding is a simple yet effective way to handle categorical variables in settings with high cardinality. The algorithm scales well to large datasets and is generalizable to a variety of common statistical learning problems. In the presented experimental results, we showed that CBM Encoding is performant on common benchmark tasks, and an attractive choice for real world datasets with regards to both accuracy and computation time. Since implementing this technique for WeWork's Lead Scoring Engine, we have experienced dramatic increases in accuracy as we are now able to generalize to rarer values of high-cardinality categorical variables. We hope to inspire further research into CBM Encoding as there are open questions around optimal priors and sampling strategies.

%
%
%

\newpage
\section{Appendix}
\subsection{CBM Encoding Algorithm}
\begin{algorithm}
\SetAlgoLined
\DontPrintSemicolon
 \caption{CBM Encoding}
 \KwData{Categorical feature matrix $\mathbf{X}$, Target variable $y$}
\KwResult{Encoded matrix, $\mathbf{Z}$}
 Initialize prior distributions $p(\theta_m)_{m}$\;
  \tcc{Build the encoding}
 \For{$m$ in $1,...,M$}{
  \tcc{Pick unique values of the feature column}
 Initialize $V = \{\mathbf{X}_{nm}\}_{n}$\;
  \For{$v$ in $V$}{
   Let $\hat{y} \equiv [y_n\ : X_{n,m} = v]_n$\;
   Compute posterior distribution $p(\theta_m|\hat{y})$\;
   Let $\mu \equiv$ first $Q$ moments of $p(\theta_m|\hat{y})$\;
   Store $f(m, v) \leftarrow \mu$\;
   }
 }
 \tcc{Build $Z$}
\For{$m$ in $1,...,M$}{
    \For{$n$ in $1,...,N$}{
        Let $\mu = f(\mathbf{X}_{nm}, m)$\;
        \For{$q$ in $1,...,Q$}{
            Set $Z_{n,(m-1)Q+q} \leftarrow \mu_q$\;
        }
    }
 }
 Return $\mathbf{Z}$
\end{algorithm}

\subsection{Datasets}
\subsubsection{Adult\protect\footnote{\url{https://archive.ics.uci.edu/ml/machine-learning-databases/adult}.} (Binary Classification)}The Adult dataset was extracted from the 1994 Census Bureau with the task of predicting whether or not someone earns more than \$50,000.  The features are a mix of numeric and categorical.

\subsubsection{Road Safety\protect\footnote{\url{https://data.gov.uk/dataset/cb7ae6f0-4be6-4935-9277-47e5ce24a11f/road-safety-data}.} (Binary Classification)}The Road Safety dataset was utilized by Cerda et. al. in 'Similarity encoding for learning with dirty categorical variables'\cite{ref_3}.  As in their paper, we randomly sampled 10,000 rows to be used in a binary classification task (target=’Sex of Driver’) with selected categorical features ‘Make’ and ‘Model’.

\subsubsection{Car Evaluation\protect\footnote{\url{https://archive.ics.uci.edu/ml/machine-learning-databases/car}.} (Multiclass Classification)}A decision framework developed by Bohanec and Rajkovic in 1990 to introduce the $\it{Decision}$ ${EXpert}$ (DEX) software package \cite{ref_21}. The model evaluates cars according to a mix of categorical and numerical features.

\subsubsection{Nursery\protect\footnote{\url{https://archive.ics.uci.edu/ml/datasets/nursery}.} (Multiclass Classification)}The Nursery dataset is a hierarchical decision model developed in the 1980's to rank nursery school applications. Features are a mix of nominal and ordinal.

\subsubsection{Insurance\protect\footnote{\url{https://github.com/stedy/Machine-Learning-with-R-datasets/blob/master/insurance.csv}.} (Regression)}The insurance dataset originates from Machine Learning with R by Brett Lantz.  The regression problem is to predict medical charges from a set of features.  The dataset is simulated from US Census demographic statistics.

\subsubsection{Bike Sharing\protect\footnote{\url{https://archive.ics.uci.edu/ml/datasets/bike+sharing+dataset}.} (Regression)}The Bike Sharing dataset contains hourly demand data from Washington D.C.'s Capital bikeshare program. The task is to predict demand from a rich feature set including weather and holidays.

\subsubsection{PetFinder\protect\footnote{\url{https://www.kaggle.com/c/petfinder-adoption-prediction}} (Multiclass Classification)}This Kaggle competition, hosted by PetFinder, asks Data Scientists to predict the speed at which animals are adopted from a mix of descriptive and online meta features. For this paper, we only use the descriptive, tabular data found in the competition's $\it{train.csv}$ file.

\subsubsection{Lead Scoring\protect\footnote{(pending legal approval) \url{https://github.com/aslakey/CBM_Encoding)}} (Binary Classification)}
The lead scoring data set is a randomly selected sample with a smaller subset of features from WeWork’s actual lead scoring data set.  The task is to predict whether or not a lead books a tour at one of our locations. We have transformed the data to preserve anonymity and mask the actual values of the features themselves.
\end{document}